\documentclass[letterpaper, 10 pt, conference]{ieeeconf}  

\IEEEoverridecommandlockouts                              

\usepackage{mathtools}
\usepackage{dsfont}
\usepackage[dvipsnames]{xcolor}
\usepackage[colorinlistoftodos]{todonotes}
\usepackage{booktabs}
\usepackage{xfrac}
\usepackage{bbm}

\usepackage{algorithm}
\usepackage{algorithmicx}

\usepackage[most]{tcolorbox}
\usepackage{xparse}
\usepackage{lipsum}
\usepackage{changepage}

\usepackage{amsmath}
\usepackage{amsfonts}
\usepackage{algpseudocode}
\usepackage{wrapfig}
\usepackage{tabularx}
\usepackage{multirow}
\usepackage{booktabs}
\usepackage{hyperref}
\newcommand{\method}{DemoStart}
\usepackage[normalem]{ulem}
\usepackage{fancyvrb}

\title{\LARGE \bf
\method{}: Demonstration-led auto-curriculum  \\ applied to sim-to-real with multi-fingered robots
}

\author{
Maria Bauza*$^1$, Jose Enrique Chen*$^1$, Valentin Dalibard*$^1$, Nimrod Gileadi*$^1$,  \\ 
Roland Hafner*$^1$, Murilo F. Martins*$^1$, Joss Moore*$^1$, Rugile Pevceviciute*$^1$, \\
Antoine Laurens$^1$, Dushyant Rao$^1$, Martina Zambelli$^1$, Martin Riedmiller$^1$,  \\
Jon Scholz$^1$, Konstantinos Bousmalis$^1$, Francesco Nori$^1$, Nicolas Heess$^1$ \\ \thanks{* Alphabetical order, equal contributions.}
\thanks{$^{1}$ Google DeepMind.}
}

\begin{document}

\maketitle
\thispagestyle{plain}
\pagestyle{plain}


\begin{figure*}[!ht]

\centering
\includegraphics[width=\linewidth]{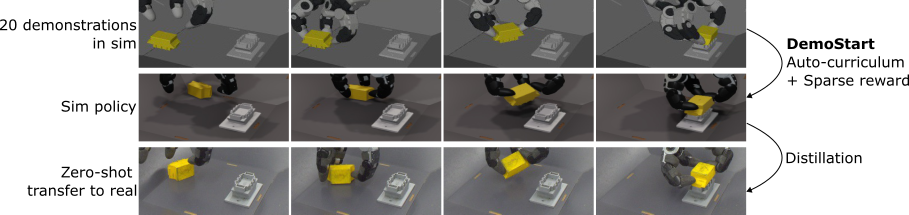}
\caption{Starting with 20 demonstrations in simulation (Top), \method{} generates a curriculum for reinforcement learning. Once a state-based policy is trained (Middle), we use distillation to turn it into a visuomotor policy that can be executed on the real robot (Bottom).}
    \label{fig:timelines}
\end{figure*}

\begin{abstract}

We present \method{}, a novel auto-curriculum reinforcement learning method capable of learning complex manipulation behaviors on an arm equipped with a three-fingered robotic hand, from only a sparse reward and a handful of demonstrations in simulation. 
%

Learning from simulation drastically reduces the development cycle of behavior generation, and domain randomization techniques are leveraged to achieve successful zero-shot sim-to-real transfer. Transferred policies are learned directly from raw pixels from multiple cameras and robot proprioception.
Our approach outperforms policies learned from demonstrations on the real robot and requires 100 times fewer demonstrations, collected in simulation.
More details and videos in 
\underline{\href{https://sites.google.com/view/demostart}{sites.google.com/view/demostart}}.

\end{abstract}

\section{Introduction}
A longstanding goal of the robotics community has been to synthesize skills that are functional, performant, and safe to execute on real robots, especially for complex robot morphologies and tasks such as manipulation with multi-fingered hands. 
Learning from human-teleoperated robots has led to some impressive results for relatively simple embodiments, for which effective teleoperation devices exist \cite{chi2023diffusion,bousmalis2024robocat,zhao2023aloha}.
However, high-quality teleoperation of sophisticated embodiments such as dexterous hands remains an open research challenge. 
As a result, simulation-based learning with subsequent transfer to real robots (sim-to-real) is an attractive alternative that can significantly reduce the scalability and safety concerns associated with learning from real robot data.

{In this work we tackle a set of problems with the following constraints: a complex embodiment with a large action space; a very small number of sub-optimal demonstrations available per task; and access to only success detectors that act as {binary sparse rewards}. Finding a general approach to solve problems of this type can greatly expand the set of tasks that can be solved with dexterous robots by reducing the amount of effort required to solve a new task.
}

Our auto-curriculum method, \method{} (Section~\ref{subsec:auto-curriculum}), combines a handful of demonstrations in simulation with RL from sparse rewards to go beyond the quality of demonstrated behaviors and avoid the need for difficult reward design. 
\method{} creates a curriculum in which the difficulty of learning is automatically adjusted. 
%
While the trained policies leverage privileged information, we distill them into policies that rely only on RGB camera images and proprioception.
In combination with the judicious use of domain randomization techniques, this allows the policies to be deployed zero-shot to real robots.
Compared to learning from demonstrations directly, our approach can learn from two orders of magnitude fewer demonstrations. 

We test \method{} on several challenging tasks that require 6D Cartesian control of an arm with 7 degrees of freedom (DoF) and joint-space control of a dexterous hand with 12 DoFs. 
We show that \method{} is effective in simulation, with over 98\% success on a number of different tasks including lifting, plug insertion, cube reorientation, nut-and-bolt threading, and a tidy-up task of placing a screwdriver in a cup. 
We also demonstrate effective zero-shot transfer of our solutions for plug lifting, plug insertion, and cube reorientation to the real world (97\%, 64\%, and 97\% success, respectively),
outperforming both the zero-shot sim-to-real transfer of naive RL baselines and policies imitating human-teleoperated demonstrations collected directly in the real world. 
Figure~\ref{fig:setup} shows the setup and tasks considered in this work. The video in the supplementary material includes further visualizations of the behaviors generated by \method{} and the baselines. In summary, the contributions of this work are: 
\begin{enumerate}
    \item The \method{} method for solving tasks in simulation from minimal and sub-optimal demonstrations, with sparse rewards.
    \item Experiments showing that \method{} can master difficult-to-control simulation tasks with a high DoF embodiment, achieving success rates over 98\% on all evaluated tasks.
    \item Successful evaluations of sim-to-real zero-shot transfer on a dexterous multi-fingered hand mounted on a robot arm, showing that \method{} generates transferable behaviors for dexterous tasks. 
\end{enumerate}

\section{Related Work}

Research in robotic manipulation is moving on from pick-and-place to more challenging dexterous tasks~\cite{kannan2023deft,aldaco2024aloha}. Along with more complex tasks, we see an increased use of bimanual robots with parallel grippers~\cite{aldaco2024aloha,bauza2023simple,varley2024embodied} and multi-fingered hands~\cite{chen2023visual,shaw2023leap}. These embodiments have higher DoFs and require more complex control.

Two approaches are common in this domain: learning policies in simulation using RL~\cite{chen2023visual, wan2023unidexgrasp,petrenko2023dexpbt,khandate2023dexterous,andrychowicz2020learning,morgan2022complex,kumar2016optimal,yin2023rotating,qi2023general}, or imitating human demonstrations on the real robot \cite{zhao2023aloha,kannan2023deft,shaw2023leap}. Combining these two approaches by kick-starting RL with demonstrations has many benefits~\cite{schaal1996learning,vecerik2017leveraging,rajeswaran-rss-18}, and allows us to use a small number of demonstrations and simple sparse rewards.

To transfer learned behaviors, some approaches rely on hand-crafted external perception mechanisms such as markers~\cite{andrychowicz2020learning,kumar2016optimal,openai2019solving}, or utilize sensor modalities such as point clouds, binary tactile sensing, or segmentation maps to minimize the sim-to-real gap~\cite{chen2023visual,yin2023rotating,qi2023general,tang2023industreal}. We operate directly on RGB images.
%
%
To transfer feature-based policies into the real world, some works first explicitly estimate the state from the RGB images~\cite{andrychowicz2020learning,morgan2022complex,openai2019solving,allshire2022transferring,handa2023dextreme}, while others rely on vision-based distillation to turn the feature-based policy into a visuomotor policy~\cite{chen2020learning}. 
We chose the latter as it doesn't rely on a fixed representation of the state space.
%

Auto-curriculum methods \cite{portelas2020automatic} have been a popular approach to train policies with RL in challenging environments \cite{paired, jiang2021prioritized, team2021open, pmlr-v202-bauer23a, schmidhuber2012powerplay}. \method{}, presented in Section \ref{subsec:auto-curriculum}, is an auto-curriculum method which has the advantage of requiring little modification to an existing RL setup and does not require a centralized controller, making it easy to scale. Prior work has also exploited the use of state resets to induce a curriculum \cite{florensa2017reverse} and used demonstrations as the source of these states \cite{popov2017dataefficient, salimans2018learning, nair2018overcoming, resnick2019backplay, tao2024reverse}. Most similar to our approach is Tao et al. \cite{tao2024reverse} which combines an auto-curriculum method, PLR \cite{jiang2021prioritized}, with state resets from demonstrations to learn policies. One key difference of our approach is that we combine the auto-curriculum and resets from demonstration into one integrated method, instead of two separate consecutive stages. We also do not feed the demonstrations as part of the replay experience, which allows us to gracefully handle demonstrations of poor quality or where actions are in the wrong action space.

\section{Method}
We propose an approach to train policies in simulation which can transfer zero-shot to real environments. Our approach requires only a few demonstrations in simulation and a binary sparse success reward, making it easy to design new tasks. Our procedure consists of two steps. We first learn a teacher policy in simulation from features (not images) using a novel auto-curriculum RL method, \method{} (see Section~\ref{subsec:auto-curriculum}), and physics domain randomization.  
We then distill the teacher policy into a vision-based student policy (Section~\ref{subsec:distillation}) and transfer it zero-shot to real. 


We consider learning within a Markov decision process (MDP) framework. At each time step, $t$, the agent selects an action, $a_t \in \mathcal{A}$, based on its current state, $s_t \in \mathcal{S}$. Subsequently, it receives a reward ($r_{t+1} = R(s_t, a_t) \in \mathbb{R}$) and transitions to the next state ($s_{t+1}$) according to the transition probability distribution ($p(\cdot \mid s_t, a_t)$). With sparse rewards, we end the episode on the first success.\looseness=-1


We parameterize the MDP with \emph{task parameters} (TPs) $\psi$. Each $\psi$ contains 1) The starting state $s_0$, 2)  Environment settings such as physical or visual parameters, and 3) A goal specification, for goal-conditioned tasks. We call $\mathcal{T}_{\text{target}}$ the distribution over TPs with starting states $s_0$ from the initial states of the target environment. In our auto-curriculum method, we extend $\mathcal{T}_{\text{target}}$ by creating TPs that are better suited for training policies. When evaluating the performance of a policy in simulation, we sample the TPs from $\mathcal{T}_{\text{target}}$.

\subsection{\method{}: A demonstration-led auto-curriculum for RL in simulation}\label{subsec:auto-curriculum}


In this section we propose \method{}, an auto-curriculum method where demonstrations are leveraged to provide increasingly difficult episode start states. \method{} leverages three distinct conceptual mechanisms which are then put together as part of a single data generation procedure. We first describe these mechanisms and then show how they fit together in \method{}.

\textbf{Conceptual mechanisms used by \method{}.} \method{} leverages the following three mechanisms:

Mechanism 1: Turning demonstrations into TPs of varying difficulty. We start by recording a few demonstrations of solving the task in simulation. For each time step in the demonstrations, we save the full environment state such that each of these states can then be used as the starting state $s_0$ of a TP. TPs that have a start state towards the end of demonstrations, and therefore require fewer steps to reach a goal state, tend to be easier, whereas TPs with start states towards the beginning of demonstrations tend to be harder.

Mechanism 2: Finding TPs with high training signal with zero-variance filtering. During learning, most TPs are either too hard or too easy to provide a good training signal for the current policy. 
For instance, if a policy knows how to insert a plug when already on top of its socket, a good TP for learning might be one where the plug is lifted and almost on top of the socket while a hard TP would be one where the plug is far from the socket and not grasped. Other parameters of a TP, such as the physics parameters, will also affect the difficulty of the task. 
\method{} uses a novel heuristic to identify TPs that lead to a strong training signal: we only train on experience generated from TPs for which the current policy sometimes succeeds and sometimes fails. We call this heuristic Zero-Variance Filtering (ZVF). 
The main intuition behind ZVF is that if success on a TP doesn't vary across episodes then there is no behavior to be reinforced in the experience of that TP. We further discuss some considerations regarding the applications of ZVF in Appendix~\ref{subsec:envstochasticity}.

Mechanism 3: Focusing on TPs with minimal bias from demonstrations. Recorded demonstrations often contain unnatural states which would not occur when executing a skilled agent.
For example, a demonstration may contain states where the agent has an unstable grasp of an object. 
As a result, training using such states as TPs' starting states can have a negative impact on performance. 
In our example, the agent may end up focusing on how to rectify a poor grasp instead of learning how to grasp an object well in the first place. 
To tackle this, \method{}'s implementation includes a novel technique to bias the TPs it trains on towards states which occur earlier in the demonstrations. This way, once an agent has become competent at grasping an object, it will no longer train on TPs where the object is already grasped. The details of how this bias is implemented is described in the overall implementation of \method{} below.

\textbf{Implementing \method{}.}
\method{} is implemented as part of a distributed actor-learner setup~\cite{espeholt2018impala}. 
The actors generate experience by executing the policy in the environment and send it to a learner through an experience replay buffer~\cite{lin1992reinforcement}. 
The learner updates the policy based on the experience using the MPO algorithm~\cite{abdolmaleki2018maximum}. 
The method we describe below determines how we gather experience for learning and thus it is implemented as part of the actors without affecting other parts of the system such as the agent update rules. Each actor repeatedly executes the following procedure: 

Step 1. Sample a sequence of TPs. First, the actor samples a sequence of TPs $(\psi_0, \psi_1, ..., \psi_K)$. 
The first TP $\psi_0$ is always sampled from $\mathcal{T}_{\text{target}}$. The next $K$  TPs are sampled by first sampling one demonstration from the set of demonstrations and splitting it into $K$ temporal chunks of equal size. We then sample uniformly a single environment state from each chunk $i$ and use it as $s_0$ of the corresponding TP $\psi_i$.
As a result, the sequence is designed so that the TPs tend to go from harder to easier. 

Step 2. Find a TP to train on. Second, the actor searches for a TP in the sequence which is identified to contain high training signal. Starting with the first TP $\phi_0$, the actor generates $T$ episodes by executing the policy on that TP. Note that none of this generated data is sent for training. We then apply the ZVF criterion to decide whether this TP has high training signal. If there is variance in the sparse reward outcomes, meaning some but not all of the $T$ episode succeeded, we consider the TP good for learning and we move on to Step 3. If there is no variance across the $T$ episodes because the agent always fails, meaning the TP is too hard, we move on to evaluating the next TP in the sequence which should be easier. If the agent always succeeds on that TP, meaning the TP is too easy, or we have run out of TPs in the sequence, we go back to Step 1 to sample a new sequence of TPs.~\looseness=-1

Step 3. Generate training data on the selected TP. Finally, if the previous step found a TP suitable for training, we then generate $M$ episodes of the policy on that TP and send the data to the replay buffer for training. When done, we return to Step 1.

We highlight three properties of \method{}. First, ZVF has the positive property that it is easy to implement as part of any setup with binary sparse rewards. Unlike other auto-curriculum methods like PAIRED \cite{paired} or PLR \cite{jiang2021prioritized}, \method{} does not require an additional agent to be trained or a separate centralized controller. Second, \method{} does not make use of the demonstrations' experience as part of the replay. As such, it is able to leverage demonstrations that are of poor quality or that are not suitable for behavior cloning, for example due to a different action space. In particular, as we show in the plug insertion task (Section \ref{subsec:tasks}), we are able to leverage demonstrations where no demonstration completes the entire task from beginning to end, but each demonstration executes a section of the task (e.g. flipping the plug to the upright position). Third, \method{} will gracefully switch training from using initial states from the demonstrations to using the initial states from the target environment in $\mathcal{T}_{\text{target}}$.
In all our experiments, we use $K=8$, $T=4$ and $M=50$.

\subsection{Distillation and Transfer}\label{subsec:distillation}

After training a policy that operates from privileged observations using \method{}, we distill the result into a student policy with visual observations, as in \cite{chen2020learning}. Instead of interactively querying the teacher policy while training the student policy, we generate an offline dataset of trajectories from the teacher, and use behavior cloning to train the student. This approach provides additional flexibility as it makes it easy to mix different types of data, or reuse the same pipeline for learning entirely from human demonstrations (see Section~\ref{sec:baselines}).

\textbf{Domain Randomization.} We use three types of domain randomization~(DR)\cite{tobin2017domain} when training the teacher and student policies: 1) Perturbations: We apply external force disturbances to any object that is not fixed to the basket; 2) Physics: We randomize the friction, mass, and inertia of every joint and body in the scene; 3) Visual: We randomize the camera poses and lighting, as well as the colors of every object in the scene (see Appendix~\ref{appendix:simulation}). When training the teacher policy with \method{}, we do not use perturbations and physics DR for episodes initialized from demonstration states.


\textbf{Photorealistic Rendering.} We include photorealistic images rendered with Filament~\cite{filament} for the plug insertion task. When generating these images, we randomize lighting and camera exposure settings. The first row of Figure~\ref{fig:setup} shows an example image.


\textbf{Policy distillation.} We use the Perceiver-Actor-Critic (PAC) model \cite{springenberg2024offline}, a scalable neural architecture designed for continuous control tasks for the student policy. While PAC supports optimizing an offline-RL objective, we solely use it for behavior cloning.
More details in Appendix~\ref{appendix:pac}.

\section{Experimental Setup}

\begin{figure*}[!ht]
    \centering
    \includegraphics[width=\linewidth]{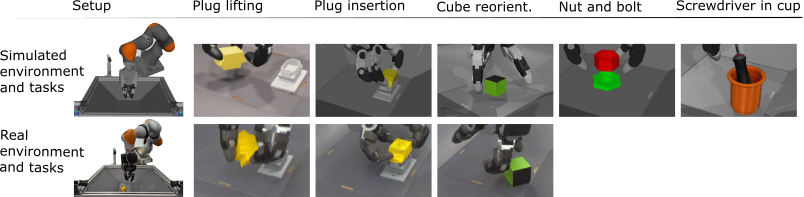}
    \caption{Experimental setup: simulated (top) and real (bottom) robot environments and tasks.}
    \label{fig:setup}
\end{figure*}

Each of our six robot cells consists of a square basket with slanted walls and two cameras fixed to the basket corners: front right and left corner. We use a Kuka LBR iiwa 14 robot arm with the three-finger \textit{DEX-EE Hand}~\cite{dex} attached to its flange, as well as two wrist cameras attached to either side of the base of the hand. The real robot setup is replicated in simulation using MuJoCo. The action space exposed to the agent is 18-dimensional. The first 6 dimensions correspond to the desired 6D Cartesian velocity of the robot arm, while the last 12 dimensions correspond to the desired joint positions of the fingers of the robot hand. Our real and simulated experimental setups are shown in Figure~\ref{fig:setup}. Please refer to Appendix~\ref{appendix:setup} for a more detailed description of the experimental setup.


\subsection{Tasks}
\label{subsec:tasks}
Our tasks are inspired by the NIST dexterous manipulation benchmark~\cite{nist_benchmark} and have varying levels of complexity, time horizon (multiple stages), precision, among other attributes relevant to dexterous manipulation (see Figure~\ref{fig:setup}).

For real robot experiments we 3D-printed a simplified version of an industrial-grade plug (yellow) and socket (gray) connector from NIST Board~\#1~\cite{nist_benchmark}, as well as a green cube with a black side. For simulation experiments we also used oversized CAD models of a nut and bolt as well as a screwdriver and cup from the YCB object dataset~\cite{Berk2017YCB}. Below is a brief description of each task, along with the type and number of demonstrations we captured from simulation for \method{}. Further details on tasks and objects in Appendix~\ref{appendix:tasks}.

\textbf{Plug lifting:} Lift the yellow plug 5cm above the basket surface. We used 5 demonstrations.

\textbf{Plug insertion:} Insert the plug into the socket. We initially recorded 12 demonstrations of plug insertion where the plug was always spawned upright. We then changed the environment to randomize the orientation of the plug and instead of collecting new demonstrations of insertion, we just added 8 demonstrations of flipping the connector into the upright position, which was faster and didn't require solving the insertion task. Our experiments use all 20 demonstrations.

\textbf{Cube reorientation:} Reorient the cube such that a specific side is facing upwards and within 5cm of the basket center. The goal orientation is sampled as part of the TP. We used 2 demonstrations of unstructured interaction with the cube.

\textbf{Nut and bolt threading:} Thread a nut onto a bolt. The nut and bolt are oversized and the bolt is fixed. We used 60 demonstrations total: 20 demonstrations of the full task, 20 lifts and 20 demonstrations where the nut was brought to the top of the bolt.

\textbf{Screwdriver in cup:} Ensure cup is upright and insert screwdriver into the cup, handle first. We used 20 demonstrations.

To collect demonstrations in simulation, we use a simpler action space that allows us to collect data through a 3Dconnexion~SpaceMouse~\cite{spacenav}. Although the action space was different, the demonstrations can be used by \method{}, which does not use the actions from the demonstrations. As part of the teleoperation baseline described in Section~\ref{sec:baselines}, we also collected expert human demonstrations on the real robot setup. Appendix~\ref{appendix:sim_teleop}-\ref{appendix:real_teleop} include further details on teleoperation.


\section{Results}\label{sec:results}
\subsection{Baselines}\label{sec:baselines}

We compare \method{} against various baselines and ablations on the plug insertion task. Both \method{} and our \textit{vanilla RL} baseline uses the same sparse reward and MPO agent architecture, as described in Appendix \ref{appendix:agentarch} (see more in Appendix~\ref{appendix:learned_behaviour}). We also compare to Scheduled Auxiliary Control (\textit{SAC-X})~\cite{riedmiller18SACX}, an RL method that uses auxiliary rewards to help exploration. In order to evaluate on the real robot, we use the distillation method from Section~\ref{subsec:distillation} to learn either from data generated from our agents in simulation or from real demonstrations obtained through teleoperation (2067 successful demonstrations for insertion and 2116 for lift). Note that for the plug insertion task, the 20 demonstrations in simulation are equivalent to half an hour of collection time. In contrast, the 2753 demonstrations collected on the real robot, including failures, are equivalent to 27 hours non-stop of data collection.
We discuss further details about our baselines in Appendix \ref{appendix:baselines}.

\subsection{Simulation experiments}

We compare the performance of agents trained in simulation on the tasks of plug lift and plug insertion. See Table~\ref{table:sim_baselines_performance} for the performance of each method on plug insertion. For each evaluation, we run 1000 episodes of 10~seconds each with different initial conditions. We include physics DR and perturbations on the objects during the evaluations. An episode is successful if the task was achieved during the episode.

Vanilla RL with the same sparse rewards was unable to solve any of the plug tasks. 
SAC-X using manually-designed auxiliary rewards can solve plug lift (100\%) and insertion (99.2\%), however this high performance comes at high cost as the process of generating the auxiliary rewards is non-trivial and requires domain expertise. \method{} also solves the tasks with 99.7\% success rate for lift and 99.6\% for insertion, but requires only a sparse reward and a few demonstrations in simulation. 
When distilling with BC, the distilled vision-based policy from \method{} data retains its performance (99.0\% for plug insertion) while distilling SAC-X in plug insertion results in a significant drop of performance (20.4\%) as the distilled policy struggles to learn from the jerky, non-smooth behaviors of SAC-X policy. Distilling \method{} and SAC-X for plug lift results in high success rates for both tasks, with 99.3\% and 98.3\% success rates respectively.
Note that the distilled policies only have access to RGB images and proprioceptive information, and their evaluations randomize the camera poses and use photorealistic images.

To evaluate the relative importance of the mechanisms introduced in Section~\ref{subsec:auto-curriculum}, we include two ablations. In \textit{Vanilla RL + Mechanism 1} actors sample the TPs from a mixture distribution. 20\% of this mixture consists of the native environment's distribution $\mathcal{T}_{\text{target}}$ and the remaining 80\% are uniformly distributed across the states from the collected demonstrations.
In \textit{Vanilla~RL + Mechanisms~1~\&~2}, each actor additionally uses the ZVF criterion to decide whether to send the experience it has generated to the learner, instead of sending all data.
In \textit{Vanilla~RL + Mechanism~1~+ Success Filter}, ZVF is replaced with a simple criterion that if any rollout from a given TP is successful, that state is considered informative.

Finally, table~\ref{table:sim_demostart_performance} shows the performance of \method{} on the five tasks in Figure~\ref{fig:setup}. \method{} achieves near-perfect results in all tasks in simulation and behavior is also significantly more efficient at solving the task compared to the demonstrations it was trained on. For instance, in the Screwdriver in cup task, agents take a median time of 3.5 seconds to solve the task, whereas the median time of the demonstrations is 93.2 seconds.

\begin{table}[ht]
\scriptsize
\centering
\begin{tabular}{| l | c | } 
\toprule
Method & Plug Insertion  \\
\midrule
Vanilla RL & 0\%  \\
Vanilla RL + Mechanism 1 & 0\%  \\
Vanilla RL + Mechanism 1 + Success Filter & 0\% \\
Vanilla RL + Mechanisms 1 \& 2 & 97.2\%  \\
\method{} (Mechanisms 1, 2 \& 3)  & 99.6\% \\ 
SAC-X & 99.2\% \\
\method{} + BC distillation &  99.0\%\\
SAC-X + BC distillation & 20.4\%  \\
\bottomrule

\end{tabular}
\caption{Performance in simulation on the Plug Insertion task.}
\vspace{-5mm}
\label{table:sim_baselines_performance}
\end{table}

\begin{table}
\scriptsize
\centering

\begin{tabular}{| l | c|}
 \toprule
 \multicolumn{2}{|c|}{\method{} (Mechanisms 1, 2 \& 3)} \\
 \midrule
 Plug Insertion & 99.6\% \\
 Plug Lift & 99.7\% \\
 Cube Reorientation & 99.9\%  \\
 Nut and Bolt & 99.8\%  \\
 Screwdriver in cup & 98.6\% \\
\bottomrule
\end{tabular}
\caption{Performance in simulation of \method{} on all tasks.}
\label{table:sim_demostart_performance}
\vspace{-10mm}
\end{table}

\begin{figure}
    \centering
    \includegraphics[width=0.45\textwidth]{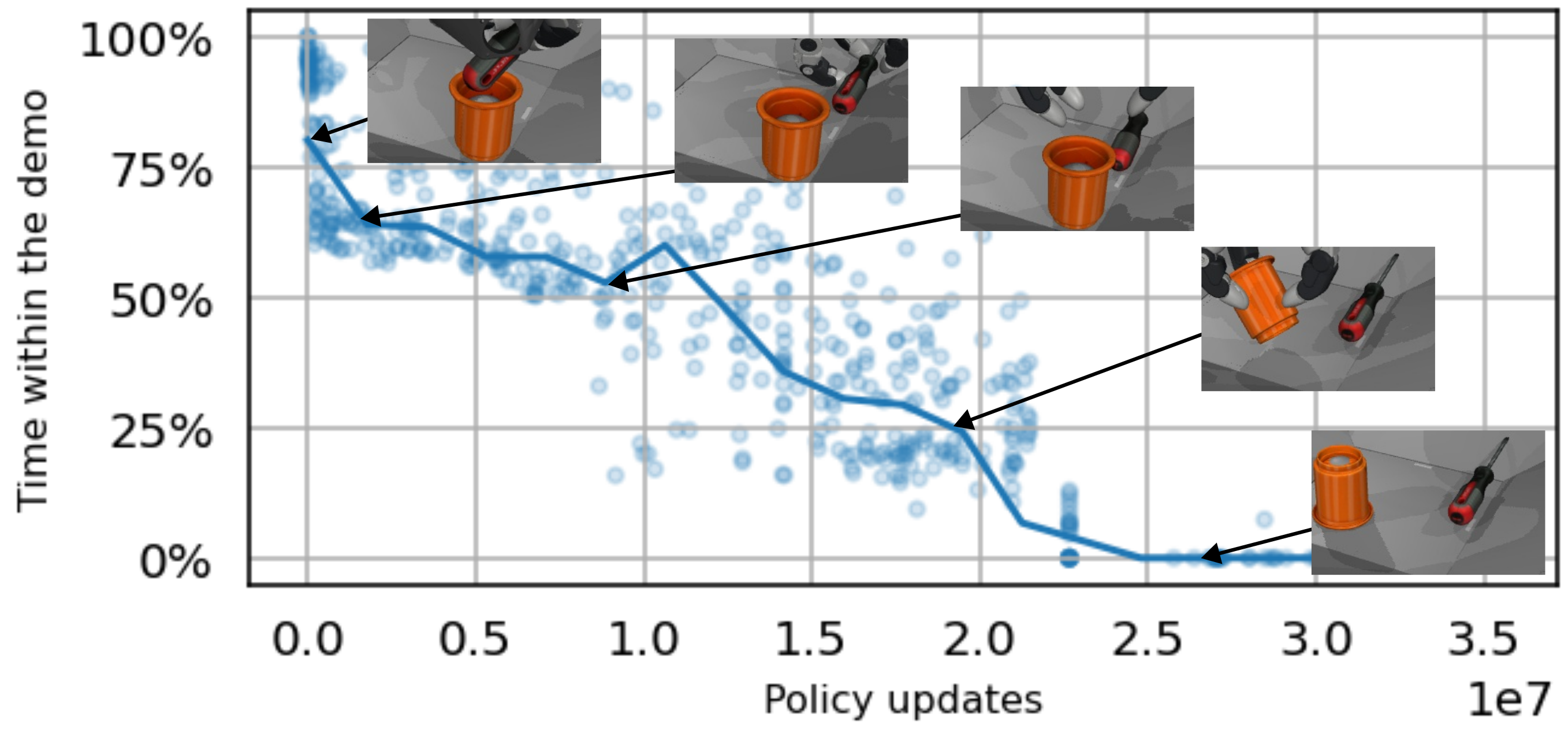}
    \caption{
    Visual representation of the emergent curriculum. We plot, for a single demonstration, the states that passed ZVF and were used for training. We see that as training progresses, the training distribution shifts from the end of the demonstration to its beginning. 
    }
    
    \label{fig:emergent}
\end{figure}

\textbf{Emergent Curricula with \method{}.} During training, we record some of the starting states that passed the ZVF. Qualitatively, we see the emergence of semantically meaningful curricula. As Figure~\ref{fig:emergent} shows, when learning the screwdriver-in-cup task, early episodes used in training start with the screwdriver above an upright cup, and later on we see the screwdriver being grasped, the cup being held, and finally episodes starting with the cup upside down.

\subsection{Real-world performance}

On the real robot, we test distilling  \method{} on three different tasks: plug lift, plug insertion and cube reorientation. For the plug tasks, we compare policies distilled from \method{} data against those distilled from SAC-X or teleoperation data. Unless otherwise specified, the distilled policies use as observations the images from the four cameras, and the joint angles for the arm and fingers as well as the arm's tool center point. For each method and task we collected 100 episodes of one minute for lift and cube reorientation, and  three minutes for insertion. We ensured that no episode started in a successful state. An episode was marked as successful if it contained at least one successful step (for plug lift, the plug needed be held for at least a second). Table~\ref{table:real_performance} shows the average success rate.

\begin{figure}
  \begin{center}
    \includegraphics[width=0.5\textwidth]{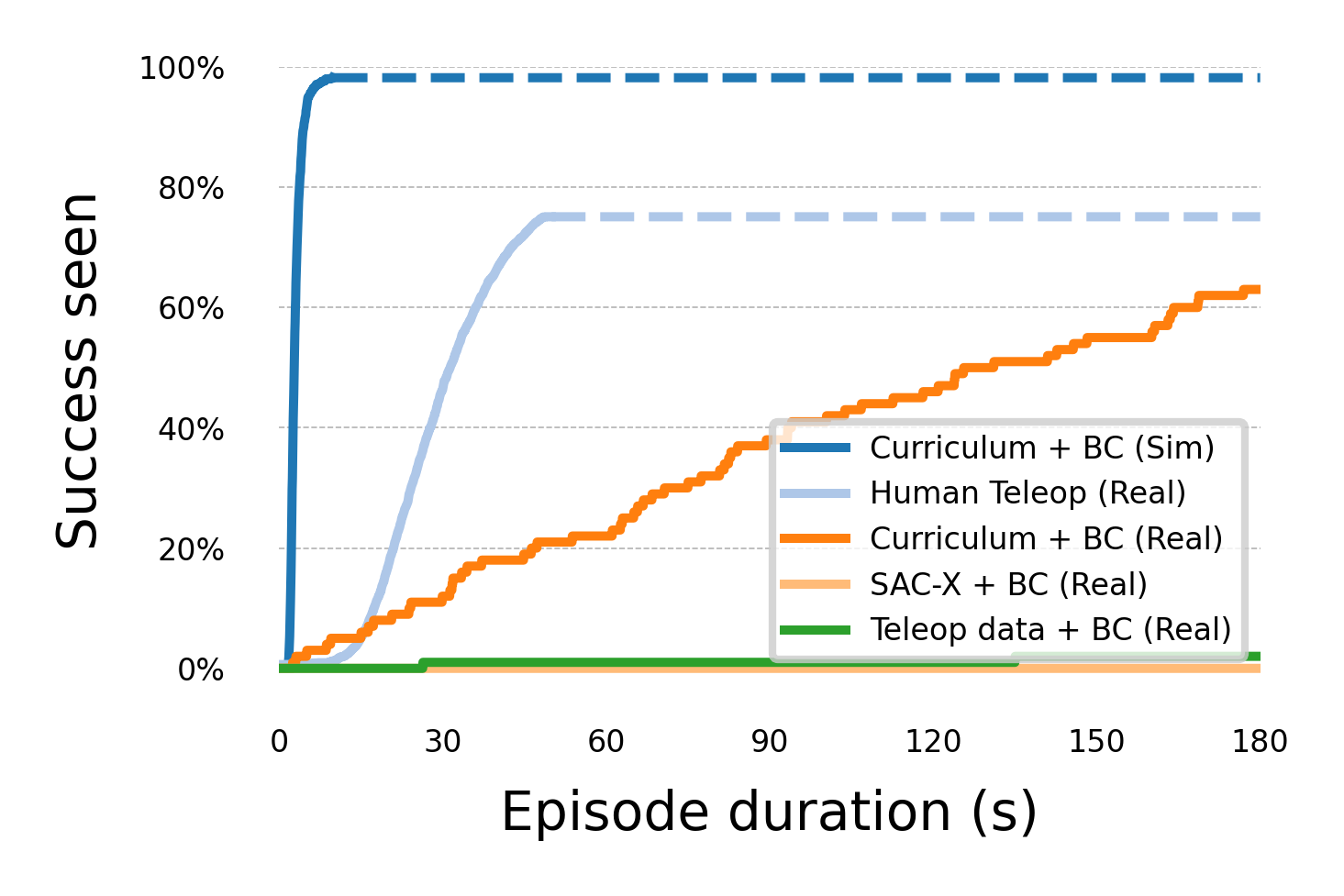}

  \end{center}
  \vspace{-7mm}
 \caption{Plug insertion success rate over episode duration. Sim episodes terminate after 10s. Teleoperation demonstrations terminate after 50s.}\label{fig:success_time}
  \vspace{-5mm}
\end{figure}

As expected, performance drops when transferring policies from simulation to the real robot. For plug lift and cube reorientation the sim-to-real gap is small and performance remains strong in the real setup with a 97\% success rate in each task. For plug insertion, the drop is not only in terms of success rate, but also in terms of how quickly the policies achieve the task. Figure~\ref{fig:success_time} shows the success rate for plug insertion as a function of episode duration, showing that in simulation policies solve plug insertion in seconds, whereas transferred policies require many more attempts.

The distilled policies from \method{} clearly outperform the other two methods on the plug tasks. It is remarkable that, although the Real Teleoperation baseline produces reasonable performance on plug lift, it fails to do so on plug insertion. Both baselines are able to capture some of the behaviors needed to solve the tasks, but result in more fidgety policies. We hypothesize these behaviors come from the action distribution in teleoperation and SAC-X data being more diverse and less smooth. For instance, each teleoperator has a different style of collecting data and will often include pauses or suboptimal motions during data collection. For SAC-X, pure RL is prone to produce jerky behaviors if not properly regularized, that in turn lead to more diverse solutions. In contrast, policies learned through  \method{} \textit{funnel} the policy behavior. This results in more consistent and smooth behaviors that are easier to learn and transfer to real. 

\begin{table}[ht]
\scriptsize
\centering
\begin{tabular}{| l | c c c | } 
\toprule
Method & Plug Lift & Plug Insertion & Cube reorientation \\
\midrule
\method{} distillation & 97\% & 64\% & 97\% \\
SAC-X distillation & 20\% &  1\% & Not evaluated \\
BC from real teleop & 64\% & 2\% & Not evaluated  \\
\bottomrule
\end{tabular}
\caption{Real robot performance of \method{} and baselines.} 
\label{table:real_performance}
\vspace{-10mm}
\end{table}

\subsection{Distillation ablations}



We perform several ablations on the distillation pipeline for learning pixel-based policies on the plug insertion task. The ablations include using the Cartesian fingertip poses as observations instead of the finger joint angles, not including photorealistic rendering in the training data, and using a different number of cameras. Table~\ref{table:sim_performance_ablations} shows the performance of PAC in the simulation and real environments for the plug insertion task. We observe that the performance in simulation is barely affected while the performance in the real world decreases significantly in some cases. For instance, the performance decreases with the number of camera views, especially when wrist cameras are not included. Not using any photorealistic data also results in lower performance. Using Cartesian poses instead of joint angles results in only a mild decrease in performance.

\begin{table}[ht]
\scriptsize
\centering
\begin{tabular}{| l | c c | } 
\toprule
Method & Plug Insertion Sim & Plug Insertion Real \\
\hline
\method{} distillation & 99.0\% & 64\% \\
\hline 
\method{} distillation & \multirow{2}{*}{97.0\%} & \multirow{2}{*}{29\%}\\
without photorealistic data & & \\
\hline
\method{} distillation & \multirow{2}{*}{98.6\%} & \multirow{2}{*}{50\%} \\
 with fingertip Cartesian poses & & \\
\hline
\method{} distillation & \multirow{2}{*}{99.3\%} & \multirow{2}{*}{51\%} \\
with 3 cameras & & \\
\hline
\method{} distillation  & \multirow{2}{*}{98.1\%} & \multirow{2}{*}{42\%} \\
with 2 cameras & & \\
\hline
\method{} distillation & \multirow{2}{*}{97.0\%} & \multirow{2}{*}{17\%} \\
 with 1 camera (no wrist camera) & & \\

\bottomrule
\end{tabular}
\caption{Ablation for the distillation pipeline settings. Unless otherwise specified, distilled policies use photorealistic data, finger joint angles and four cameras.}
\label{table:sim_performance_ablations}
\vspace{-10mm}
\end{table}

\section{Discussion}
We introduce a pipeline that leverages simulation-based training and sim-to-real transfer to synthesize complex behaviors for real robot manipulation. \method{} simplifies task design, relying only on a handful of low-quality demonstrations and a simple, sparse reward function. \method{} can also leverage incomplete demonstrations of a task and combine them in order to learn to solve the task end-to-end.
We have shown that \method{} is effective on a number of challenging tasks and that, with distillation, it can produce vision based controllers that can perform well on real robots.

Inherently, \method{} makes a number of trade-offs.
Using demonstrations only to shape the initial state distribution of RL allows greater flexibility in how the demonstrations can be obtained while allowing for more efficient and performant behaviors to emerge.
However, this simplicity can result in behaviors that don't follow the demonstrations closely and thus it is harder to provide guarantees with regards to the final solution. Another consideration to make is that, although we found the variance-based signal of \method{} to be generally effective, there are some situations where it can be misleading. This is primarily due to inherent stochasticity of the environment which can cause variance in situations where there is little training signal for the policy to learn.

Overall we expect \method{} to work well for many manipulation tasks with predictable dynamics including those with challenging exploration or hard-to-define rewards.
Regarding the limitations of this work, we believe that including more informative rewards, when available, will allow for more precise behavior shaping. Additionally, employing more advanced randomization techniques may enable better sim-to-real transfer.
Because zero-variance filtering drops most of the data produced by actors, and because we are using sparse rewards, the method is compute intensive compared to methods that use dense rewards. Compute efficiency can be improved by reducing the cost of simulation (e.g. by using a GPU-based simulator), or by better selection of informative states from demonstrations (perhaps by estimating the success probability of a starting state before rolling out).
We leave these explorations for future work. On the potential extensions, we believe the modularity of our pipeline is an asset. One interesting direction to be explored in future work, for instance, is to include real-robot data in the second training phase. This is already enabled by the current distillation pipeline and would allow for a self-improvement loop where performance continues to improve during deployment.

\bibliographystyle{IEEEtran}
\bibliography{references}  

\clearpage
\newpage
\section{Appendix}

\subsection{Agent training details}
\subsubsection{\method{} agent architecture and reward}~\label{appendix:agentarch}


\textbf{Agent architecture.} We now describe the architecture of the agent used by \method{} and in our ablations. The agent is an MPO~\cite{abdolmaleki2018maximum} agent with a shared encoder between the actor and the critic.

\begin{table}[ht]
\scriptsize
\centering
\begin{tabular}{| l | l | l | } 
\toprule
Observation name (dimensions)  & Description \\
\midrule
iiwa14\_joint\_pos (7)&  Joints position of the Kuka arm \\
iiwa14\_joint\_vel (7)& Joints velocity of the Kuka arm \\
iiwa14\_tcp\_pos (3) & Arm Tool Center Point~(TCP) position \\
iiwa14\_tcp\_pose (7) & Arm TCP position and quaternion \\
iiwa14\_tcp\_rmat (9) & Arm TCP rotation matrix \\
finger\{$0 \vert 1 \vert 2$\}\_ftip\_pos (3)& Hand fingertip position\\
finger\{$0 \vert 1 \vert 2$\}\_ftip\_rmat (9)& Hand fingertip rotation matrix\\
finger\{$0 \vert 1 \vert 2$\}\_joint\_pos (4)& Hand finger joints positions\\
finger\{$0 \vert 1 \vert 2$\}\_joint\_torques (4) & Hand finger joints torques\\
finger\{$0 \vert 1 \vert 2$\}\_joint\_vel (4)  & Hand finger joints velocities\\
finger\{$0 \vert 1 \vert 2$\}\_joint\_commanded (4) &  Commanded finger joint positions\\
\bottomrule

\end{tabular}
\caption{Proprioceptive observations of \method{} agents}
\label{table:agent_obs}
\end{table}

The list of observations received at each time step by the agent is listed in Table \ref{table:agent_obs}. To process these observations into a policy, we first stack the observations from the last three timesteps. These vectors are then flattened and concatenated, linearly projected as a 512 vector, and passed through a Layer Normalization layer \cite{layernorm}. The shared encoder processes the normalized observation vector using an 8 layer MLP with hidden size 512. The resulting embedding is then processed by a linear policy head.

\textbf{Agent action specifications.} Two action specifications are available to control the fingers of the robot hand. An agent can use  \emph{integrated joint velocity} control in which case it outputs velocities for each of the joints, or \emph{joint position} in which case the agent outputs the target position of each of the joints directly. We found that it was beneficial to tune the choice of action specification as a hyperparameter on each task. As a result, we use joint position on all tasks apart from cube reorientation where we use joint velocity.

\textbf{Agent reward.} \method{} is able to leverage sparse rewards. In each of our tasks, if the agent reaches a goal state where the task has been completed, we terminate the episode and give the agent a reward of 1. Below we include additional details about the rewards used for the plug lifting task and the nut and bolt threading task.

In the plug lifting task, we found that having a reward for lifting the plug 5cm above the basket surface led to the agent flicking the plug up in an unstable way. We therefore changed the reward so the agent had to hold the plug 5cm above the basket surface for a consecutive 0.5 seconds before scoring a reward and terminating the episode.

In the nut and bolt threading task, we initially found that while the agent could learn to thread a nut already placed on top of the bolt, it would fail to learn to grasp a nut and place it on top of the bolt. We found a simple modification that let the agent train successfully: we added an extra parameter \texttt{difficulty} to the task parameters $\psi$ for that task which could take the values \texttt{easy} or \texttt{hard}. If \texttt{difficulty} was \texttt{hard} the agent would receive a reward when the task was completed, with the nut fully screwed onto the bolt, as usual. If \texttt{difficulty} was \texttt{easy}, the agent would receive a reward and the episode would terminate if the nut was positioned anywhere vertically above the bolt. We found randomly sampling \texttt{difficulty} to be \texttt{easy} with a 50\% chance when generating a task parameter encouraged the agent to learn to grasp the nut and bring it to the bolt and was enough to steer the agent into completing the whole task. 

\textbf{Training hyperparameters.} Our agents were trained on trajectories of length 10. The replay buffer was configured to use each trajectory twice for training. We trained our agents with a batch size of 256. Agents were trained until convergence which typically took under 10M learner updates but took 32M learner updates on the Screwdriver in cup task. In all our ablations in Table \ref{table:sim_baselines_performance}, the agents were trained for twice the number of update steps as \method{}.

\subsubsection{Perceiver-Actor-Critic (PAC) model}
\label{appendix:pac}
In this work, we use the Perceiver-Actor-Critic (PAC) model \cite{springenberg2024offline} to distill the policies from \method{}. PAC is a scalable neural architecture specifically designed for continuous control tasks for the student policy. While PAC supports optimizing an offline-RL objective, we solely use it for behavior cloning.

To handle multimodal inputs, the model first encodes the inputs into embedding vectors using modality-specific encoders for vision and proprioception. 
$
e_I = \phi_P(s_t^P) \oplus \phi_V(s_t^V) \oplus \phi_V (\tau_V) \oplus \phi_L(\tau_L) \in \mathbb{R}^{N \times D_I}.
$ 
To reduce computational complexity, a cross-attention block is used to query the input embeddings with a small number of trainable latent vectors. 
This is followed by several self-attention operations on the latent vectors to produce the final output ($z_M \in \mathbb{R}^{N_Z \times D_Z}$).
This perceiver-like architecture reduces the computational and memory usage of self-attention by shifting it from the input embeddings to the latent vectors.
PAC policies are trained to effectively adapt to the inevitable discrepancies in dynamics, physics (such as friction at fingertips), and visual inputs that arise across different source of robotic data.

The PAC model was chosen in this paper as it provides an effective approach for learning policies from large and complex datasets while keeping the computational cost manageable. 

\subsubsection{Data and observations for the distillation pipeline}

Unless otherwise specified, the distillation policies take as input the four cameras from the cells (two on the wrist of the robot and two on the basket) downsized to squares of 80x80 pixels. The policies also take as input proprioception information: arm joint positions (seven dimensions), arm tool center point position and orientation (twelve dimensions using the rotation matrix) and the joint angles of each finger (twelve dimensions, four per finger). 

We generate the data to train the distillation policies by saving successful episodes of the feature-based policies generated by either \method{} or SAC-X. When generating the data, we include physics DR, object perturbations and visual DR. For the connector insertion task, we also include episodes with photorealistic images. For each task we generate thousands of successful episodes and when using photorealistic data, the ratio is around one episode with photorealistic rendering per one of visual DR.

\subsection{Experimental Setup}\label{appendix:setup}


We carefully designed our experimental setup with flexibility in mind, allowing for experimentation on a variety of tasks. In this study our focus is primarily on dexterous manipulation, covering dexterity attributes such as independent multi-finger control, making and breaking contacts, speed and dynamics, precision and exploitation of environmental constraints.

\subsubsection{Standard robot cell}
Our robot is placed in a cell
consisting of a basket with slanted walls that determine the workspace where tasks are carried out. We use two Basler ACA1920-40GC GigE cameras with standard lenses. The cameras are installed on front-left and front-right corners of the basket. We use 7 DoF Kuka LBR iiwa 14 R820 robot arms with the DEX-EE~\cite{dex} robot hands attached to their flange. We also attach two Basler ACA1920-40UC USB3 cameras with standard lenses to either side of the base of the DEX-EE Hand. Figure~\ref{fig:setup} includes a visualization of the setup. 

\begin{figure}
    \centering
    \includegraphics[width=8cm]{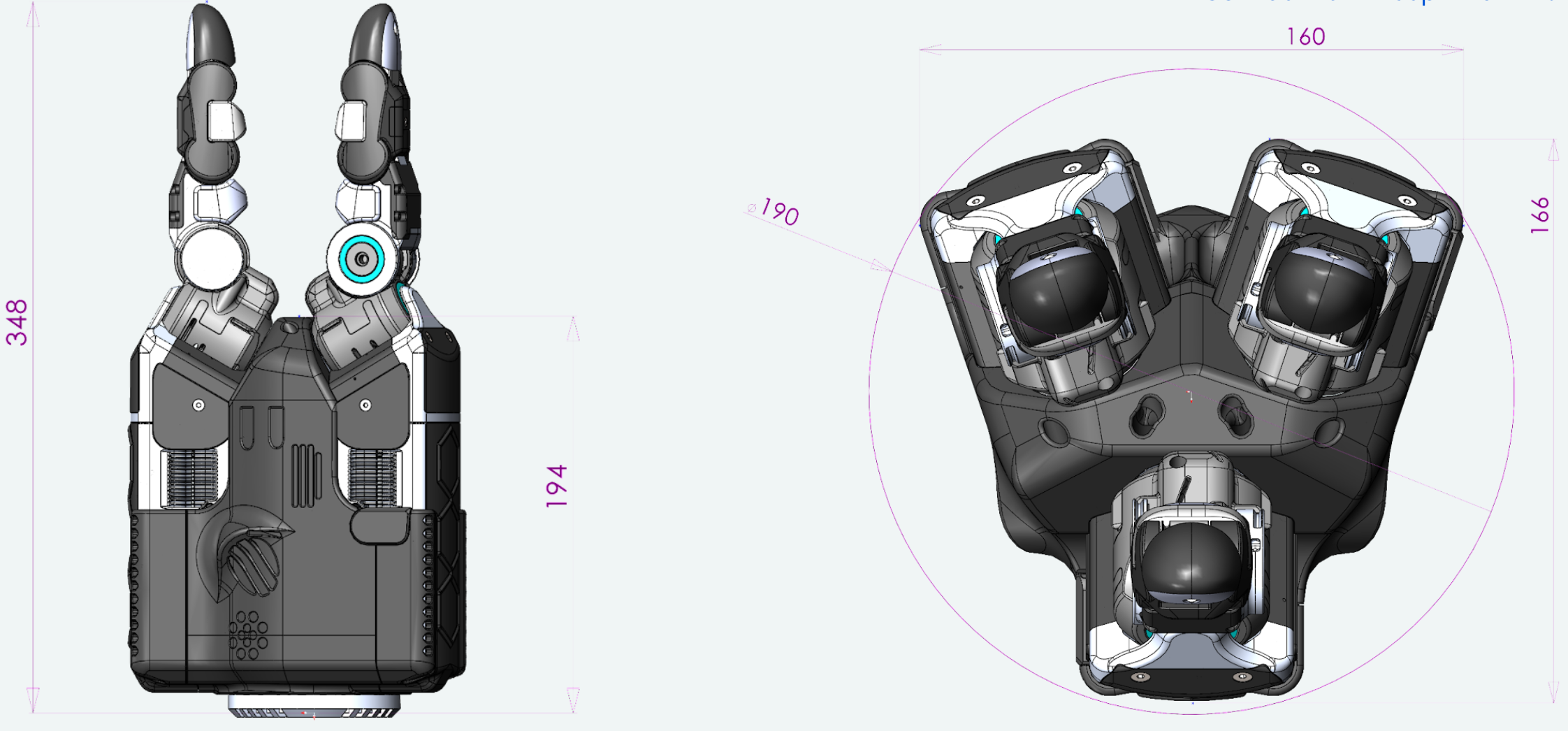}
    \caption{DEX-EE Hand - finger configuration (dimensions in mm).}
    \label{fig:appendix_dex_config1}
\end{figure}

The DEX-EE Hand~\cite{dex} 
is an over-actuated tendon-driven robot hand with 3 fingers, each of which with 4 joints and 5 actuators, with a total of 12~DoFs. The fingers are arranged in a near antagonistic configuration, with one finger opposing the other two fingers, which are almost parallel to each other~(Figure~\ref{fig:appendix_dex_config1}).

In order to maximize the workspace and manipulability of the arm, we mounted the arm on a base that is off-center and rotated w.r.t. the basket as shown in Figure~\ref{fig:setup}. We made use of a fleet of 6 real robot setups for this study.


\subsubsection{Sensing and control}

We deliberately make use of asynchronous data streams; we believe that methods that are capable of handling asynchronous data streams are more flexible, robust and much easier to deploy when compared to the non-trivial amount of work and cost to synchronise data acquisition. Several concurrent processes run in the background to fetch data from sensors and compute control commands to the robots while ensuring safety, all asynchronously. The basket cameras operate at 42 frames per second~(FPS), while wrist cameras operate at 30~FPS. The sensor data rate of both the robot arm and the robot hand is 1~kHz.

The agent and environment run in lockstep at 20~Hz, where at each step the environment will fetch the latest readings from the different data streams, and accept new commands from the agent. The robot arm and robot hand are controlled independently. The robot arm is operated by the agent with a 6D Cartesian velocity controller. The 6D Cartesian velocities commanded by the agent are first mapped into joint velocities through a QP-based differential inverse kinematics controller running in a separate process at 200~Hz. We subsequently compute adequate 7-dimensional joint torques for the robot arm that track the joint velocity setpoints through an impedance controller running at 1~kHz. The QP-based controller also incorporates a kinematic safety layer to prevent the robot arm from colliding with the basket and cameras, while allowing the robot hand to physically interact with the basket surface. The robot hand is operated with a joint position controller running in a separate process at 1~kHz.


The action space exposed to the agent is 18-dimensional. The first 6 dimensions correspond to the desired 6D Cartesian velocity of the robot arm, while the last 12 dimensions correspond to the desired joint positions of the fingers of the robot hand.

\subsubsection{Simulation}\label{appendix:simulation}

We leverage the scalable characteristics of simulation with our method. A \textit{digital twin} equivalent of the real robot setup was created using the MuJoCo simulator~\cite{todorov2012mujoco}. We use the MuJoCo model for the Kuka LBR iiwa 14 from the MuJoCo~Menagerie~\cite{menagerie2022github}.


Note that in contrast with the real robot setup where different sensors resulted in asynchronous data streams and data aliasing, in simulation all sensing data arrives synchronised as part of the physics state. Similarly, control commands are synchronously applied at every simulation step, whereas on the real robot we observe thread concurrence and network communication delays between different machines. Please refer to a related discussion in the domain randomisation section~\ref{subsec:distillation}.

As well as the default rendering provided by MuJoCo, we developed a custom solution for rendering high-quality images of the plug insertion scene, which we use to evaluate how photorealistic renderings of simulated camera images impact performance of the learning agents and the sim-to-real transfer. For this, we used the Filament renderer~\cite{filament} with custom 3D assets, materials and lighting built using Blender \cite{blender}. Sample images from the photorealistic renderer are shown in the second row of Figure~\ref{fig:timelines}.

\begin{figure*}
    \centering
    \includegraphics[width=12cm]{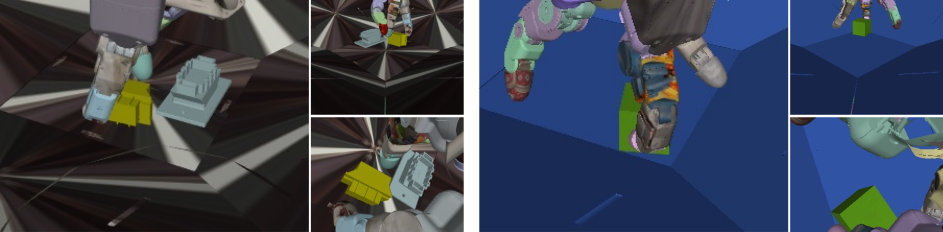}
    \caption{Example images of domain randomization.}
    \label{fig:appendix_DR}
\end{figure*}

We also use domain randomization, as explained in Section~\ref{subsec:distillation}. Examples of visual domain randomization are shown in Figure~\ref{fig:appendix_DR}).

\subsubsection{Teleoperation for simulation demonstrations}\label{appendix:sim_teleop}

To simplify data collection in simulation, we used an environment where the hand was controlled with finger joint velocity commands instead of joint position commands. Note that although the action space of the environments was different, the demonstrations could be re-used for our auto-curriculum approach, which does not use actions from the demonstrations.


Simulated robot demonstrations were collected using a 3DConnexion SpaceMouse~\cite{spacenav}. The SpaceMouse axes were used to control the 6D Cartesian velocity of the arm, while the buttons and keyboard were used to control the hand through a 8~DoF abstraction command to control the fingers of the hand. 

\subsubsection{Teleoperation for real robot demonstrations}\label{appendix:real_teleop}

Real robot demonstrations were collected using two devices, one for the arm and one for the hand. To control the arm, we used a custom magnetic-based tracking device to measure the Cartesian 6D pose of a remote controller relative to a receiver. At every step, we computed a Cartesian 6D velocity action for the arm through a P-controller on the SE(3) error between the current robot hand pose and the measured hand pose of the human teleoperator. The hand was directly controlled using the Shadow Hand Glove~\cite{glove} and a custom IK solution to compute the joint positions for the fingers.

To collect expert human demonstrations on the real robot for the baselines, a team of 6 individuals with no relation to the research team attempted the tasks on-site. For the plug insertion task, the 20 demonstrations in simulation are equivalent to half an hour of collection time, while the 2753 real demonstrations, including failures, are equivalent to 27 hours. The real demonstrations were collected over a span of 36 days as it is uncomfortable to teleoperate with the glove for over one hour.


For the lifting demonstrations we found that 500 successful demonstrations were not sufficient to achieve good performance, so we included the lift part of the plug insertion demos to get over 2000 demonstrations for plug lifting. 

\subsection{Tasks}
\label{appendix:tasks}

We take inspiration from the dexterous manipulation benchmark created by NIST~\cite{nist_benchmark}. With a focus on Task Board~\#1, we created 3D-printed versions of the objects of one of the industrial-standard connectors and nuts and bolts. By parameterizing and 3D-printing these objects, we can easily vary the scaling of such objects, create a range of plugs with different mass densities, fitting tolerances (for plug insertion) and thread angle (for nut and bolt threading). These parameterised objects also facilitate the development of simulation models with domain randomization in mind. See Figure~\ref{fig:appendix_plug_and_socket} for an example of the 3D-printed plug and socket objects.

The tasks chosen for this study varying levels of complexity, time horizon (multiple steps), precision, among other attributes relevant to dexterous manipulation. For all tasks, at the beginning of each episode, the robots are randomly moved to a given set of joint positions uniformly sampled from a predefined distribution that lies within the operational workspace of the robots. Similarly, in simulation objects are also randomly placed within the basket by uniformly sampling their initial position from a predefined distribution. For real robot evaluations we rely both on the inherent variations of the physical interactions between robots and objects during episodes, and on the regular monitoring of evaluations to ensure a wider distribution of initial states is covered.


\subsubsection{Plug lifting}

Before an agent is able to perform more dexterous tasks, one of the first behaviors that must be learned is that of lifting an object. We use or 3D-printed yellow plug (Figure~\ref{fig:appendix_plug_and_socket}). The goal of the task is straightforward: the agent must control the robots such that the plug is grasped and then lifted up from the basket surface.In this task the grey socket is present, fixed to the basket surface, but merely serves as a distractor for the agent.

The agent receives a reward of 1 if the plug center of mass is lifted 11cm or more above the basket surface for at least 0.5 seconds. i.e., the following function would have to evaluate to 1 for 0.5 seconds:

\begin{BVerbatim}[fontsize=\small]

def reward(plug_position):
  lifted = plug_position[2] > 0.11
  centered = all(
      abs(plug_position[:2]) < 0.1)
  if lifted and centered:
    return 1.0
  return 0.0
  
\end{BVerbatim}

\subsubsection{Plug insertion}

Inspired by the NIST benchmark~\cite{nist_benchmark}, this task uses our 3D-printed yellow plug and grey socket objects. These objects are designed with two symmetrical axes, increasing the level of difficulty the agent must overcome from a perception perspective when estimating their orientation from raw pixels in our real robot evaluations.

\begin{figure}
    \centering
    \includegraphics[width=8cm]{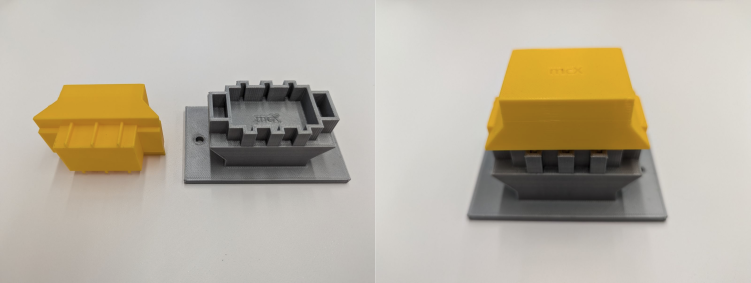}
    \caption{Our NIST-inspired 3D-printed plug and socket used in the experiments.}
    \label{fig:appendix_plug_and_socket}
\end{figure}

The goal of the task is to insert the yellow plug into the socket as illustrated in the right-hand image of Figure~\ref{fig:appendix_plug_and_socket}. The position of the socket (permanently fixed to the basket) does not change across episodes. Differently from the plug lifting task, here the agent must learn emergent behavior that grasps and reorients the plug (in-hand or exploiting environmental constraints such as the basket geometry), as well as fine motor control to precisely insert the plug into the socket.

The agent receives a reward of 1 if the plug is within 2mm of the fully inserted position, which is only possible if it's correctly oriented.

\begin{BVerbatim}[fontsize=\small]

def reward(plug_position):
  error = plug_position - SOCKET_POSITION
  if norm(error) <= 0.002:
    return 1.0
  return 0.0

\end{BVerbatim}

\subsubsection{Nut and bolt threading}

Also taking inspiration from the NIST benchmark~\cite{nist_benchmark}, we create a nut and bolt threading task in simulation. Increasing the level of difficulty from the plug insertion, in this task the agent must not only learn how to grasp, re-orient and insert the nut onto the bolt, but must also apply repeated turnings to move the nut down the bolt shaft. Similarly to the plug insertion task, the bolt is fixed to the basket surface facing upwards. The bolt position does not change across episodes.

At evaluation, the agent receives a reward of 1 if the nut is threaded onto the bolt such that the top of the nut is flush with the bolt. During training, we mix this with an easier reward as discussed in Appendix \ref{appendix:agentarch}.

\begin{BVerbatim}[fontsize=\small]

def reward(nut_position, difficulty):
  # difficulty is a value 
  # from 0 (easy) to 1 (hard).
  error = (nut_position - 
           FULLY_THREADED_POSITION)
  z_threshold = (1 - difficulty) * (
      TOP_OF_BOLT_Z - 
      FULLY_THREADED_POSITION[2])
  if (norm(error[:2]) <= 0.01 and 
      error[2] <= z_threshold):
    return 1.0
  return 0.0

\end{BVerbatim}

\subsubsection{Screwdriver in cup}

Both plug insertion and nut and bolt threading present challenging insertion problems the agent must address, with different contact interactions, multiple steps to complete the tasks and precise motions. One caveat of both tasks is that both the socket and the bolt are rigidly attached to the basket surface and their placement does not change over the course of experiments and evaluations.

We devise a screwdriver in cup task, where we selected the two objects (screwdriver and cup) from the well-known YCB object dataset~\cite{Berk2017YCB}. This task addresses the issue of two free moving objects within the basket, relaxes the requirements for precise insertion while adding requirements for gentle contact interactions (i.e. to avoid tipping the cup) and increases the time horizon of the task solution. The poses of screwdriver and cup are randomly sampled from predefined uniform distributions at the beginning of each episode. 

The agent receives a reward of 1 when the screwdriver is located upside-down inside the cup, but only if the cup is standing upright. The agent must learn to firstly ensure that the cup is upright within the basket, and then place the screwdriver upside-down inside the cup. With the random initial pose of the objects, the agent must learn grasping, re-orientation and placement of \textit{both} objects and handle contact forces and interactions between the two free moving objects.

\begin{BVerbatim}[fontsize=\small]

def reward(screwdriver_handle_position, 
           cup_position, 
           cup_z_axis):
  # up to 10 degree error
  cup_upright = cup_z_axis[2] >= 0.985  
  if (cup_upright and 
      norm(screwdriver_handle_position - 
           cup_position) <= 0.01):
    return 1.0
  return 0.0

\end{BVerbatim}

\subsubsection{Cube reorientation}

We also experimented with a cube reorientation task in simulation. The goal of this task is for the cube to be reoriented to have a specific side facing upwards when resting on the basket surface. The target pose (which of the 6 sides should face upwards) is randomly sampled from a uniform distribution. This task allows us to evaluate how the agent performs when independent control of multiple fingers is required and a relatively accurate estimation of object orientation is pivotal for solving the task. It also demonstrates the training of goal-conditioned policies using the auto-curriculum method.

The goal is specified as an axis in the object coordinates that should face upwards.

\begin{BVerbatim}[fontsize=\small]

def reward(cube_position, 
           cube_orientation, 
           goal_z_axis):
  rotated_goal = cube_rotation @ goal_z_axis
  # return 1.0 if within 8 degrees of 
  # correct alignment, and near the basket 
  # center.
  if (rotated_goal[2] > 0.99 and 
      norm(cube_position) < 0.05):
    return 1.0
  return 0.0

\end{BVerbatim}


\subsection{Baselines}\label{appendix:baselines}

\subsubsection{Vanilla RL}

Our Vanilla RL ablations presented in Table \ref{table:sim_baselines_performance} all use the same agent architecture and reward as \method{} presented in Appendix \ref{appendix:agentarch}.

\subsubsection{Scheduled Auxiliary Control}
\label{subsec:expsacx}



We apply Scheduled Auxiliary Control (SAC-X)\cite{riedmiller18SACX} to our domain. SAC-X is method that overcomes the exploration problem by defining auxiliary tasks that do not limit the final solution, which still is learned on the (potentially sparse) original task reward. As discussed below, finding good solutions to long-horizon sparse tasks requires a considerable amount of engineering in designing these auxiliary reward functions and regularising reward components.


Following the SAC-X paper~\cite{riedmiller18SACX} we define by trial and error a set of auxiliaries that allows the agent to discover rewards for the sparse task reward.
This set of auxiliaries can be built by simple to define sparse auxiliary rewards that guide the exploration and help the policy head responsible for the main task to learn with the original task reward.

\begin{figure}
    \centering
    \includegraphics[height=3.5cm]{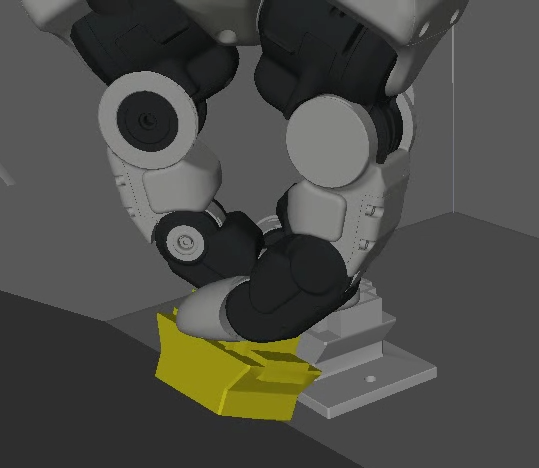}
    \includegraphics[height=3.5cm]{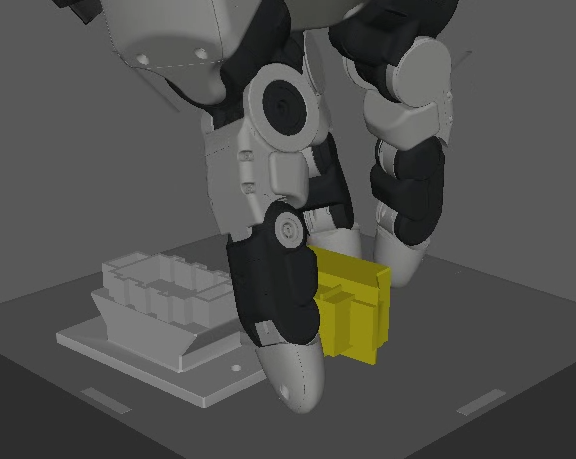}
    \caption{Example of a non regularised reach behavior (left) versus a pose regulated reach behavior (right).}
    \label{fig:appendix_reachraw}
\end{figure}

In the original paper, this was shown to work for a robotics manipulation task with a parallel gripper. A required auxiliary used there was "reaching" an object.
However, if we simply apply the same reward in this setting, the high dimensional nature of the multi-fingered hand poses additional issues.
For example a simple reaching behavior will learn to reach the object under control, but it will not care about the positions of the fingers and these end up in positions that do not facilitate manipulating objects (an example is shown in Figure~\ref{fig:appendix_reachraw} on the left).

The solution is to move away from the sparse auxiliary definitions and add regularisation on top of them. We found that a necessary regularisation is to have the hand in an open configuration with the fingers being all in a straight configuration, which we call a "comfy" base pose (an example is shown in Figure~\ref{fig:appendix_reachraw} on the right).
We also found that using this regularisation is only necessary for the reaching and lifting auxiliaries, as later auxiliaries will be biased enough to find better configurations.
In addition we found that learning with sparse and dense auxiliaries helps to learn better policies.

\begin{table}[h]
\scriptsize
\centering
\begin{tabular}{| l | l | } 
\toprule
auxiliary & description \\
\midrule
REACH COMFY & regularised, dense reach reward\\
REACHED COMFY & regularised, sparse reach reward\\
LIFT COMFY & regularised, dense lift reward\\
LIFTED COMFY & regularised, sparse lift reward\\
HOVER & dense positioning reward\\
INSERT & dense insertion reward based on positioning\\
INSERTED (task reward) & sparse insertion reward\\
\bottomrule

\end{tabular}
\caption{Example of the auxiliary rewards used by the baseline SAC-X to solve the two plug tasks. For the plug lifting task we use only the first four of the auxiliaries, while we use the full list for the insertion task. }
\label{table:appendix_sacx_auxiliaries}
\end{table}

If we apply SAC-X with the defined auxiliaries on the two plug tasks, we can make an MPO agent, similar to the one presented in Section \ref{appendix:agentarch}, and solve the tasks in simulation.
We use the SAC-Q method here where we put the task reward as the main auxiliary in SAC-Q.

The learning starts with random initialized networks and an empty transition data buffer, demonstrations are not necessary and not used in this setting.
While we do not optimize for data efficiency, as we only run these rewards in simulation in the sim-to-real setting, we need approximately 800k episodes to learn the lift task and 2m episodes to learn the insertion task from scratch.

\subsubsection{Learning from human demonstrations on the real robot}\label{subsec:human_demos}

Following the procedure in Section~\ref{appendix:real_teleop}, we collected human-teleoperated expert data for for plug lifting and plug insertion. Each episode was limited to 50 seconds, during which the teleoperators were asked to complete the task and signal whether the episode was a success or a failure. At the end of the episode, the teleoperators were asked to manually randomize the plug starting pose before the next episode started. Table~\ref{table:appendix_human_demos} shows a summary of the data collected. 

\begin{table}[ht]
\scriptsize
\centering
\begin{tabular}{| l | c c c| } 
\toprule
Task & Successful Episodes & Total Episodes & Success Rate \\
\midrule
Plug Lift &  556 & 556  & 100\% \\
Plug Insertion &  2067 & 2753  & 73\% \\
\bottomrule
\end{tabular}
\caption{Human-teleoperated expert data in real.}
\vspace{-0.5cm}
\label{table:appendix_human_demos}
\end{table}

While inspecting the episodes we found that human-to-human variability resulted in very different behaviors when solving the plug insertion task. For example, some teleoperators preferred to lift the connector and drop it to re-orient it, while others preferred using in-hand re-orientation techniques to re-orient the plug. There was also a large diversity in initial states, as we encouraged them to reset the plug in challenging or difficult to reach positions. Plug lift was a relatively simple task for the teleoperators, achieving a 100\% success rate. On the other hand, plug insertion was significantly more difficult, achieving only a 73\% success rate, as it required re-orienting the object first and carefully controlling the robot to insert the plug in the socket.




\subsection{Handling Environment Stochasticity}

\label{subsec:envstochasticity}

Zero-variance filtering works best in regimes where there is low environment stochasticity. In this context, if there is variance in the reward, it is likely due to the difference in the actions sampled by the policy, which indicates that there is learning signal as some actions should be reinforced. However, if the environment is stochastic, the variance in reward can be a result of the environment's stochasticity. In this case, ZVF may diagnose a task parameter $\psi$ to be good for learning regardless of the policy's behavior.

For example, in one of the demonstrations of the cube reorientation task, the cube was lifted and dropped from a height. Just like rolling dice, even with deterministic simulation, the result of episodes that start just before the cube left the hand was highly random, and dependent on small action perturbations. This means that even late into the training process, data from this state would pass the zero-variance filter.

In general, when possible it is preferable to generate demonstrations which only contains stable states to alleviate this issue. The specific demonstration above was removed from our train set and instead we used a demonstration with a stable grasp throughout the episode. However, we also subsequently designed Mechanism 3, presented in Section \ref{subsec:auto-curriculum}, to bias the training away from such states in the middle of demonstrations. We found this improved stability, convergence speed as well as final performance.

\subsection{Learned Behaviors}\label{appendix:learned_behaviour}

\subsubsection{Moving away from demonstrations}

The policies learned with \method{} are able to extract useful behaviors from demonstrations, such as pushing the plug down to ensure it is connected (see an example in the supplementary video). However, they can also move away from these behaviors. We find that the curriculum policies are substantially faster at solving the tasks: our policy for the task of screwdriver in cup takes about 3.5 seconds to solve the task, whereas the median time of the demonstrations is 93.2 seconds. Beyond becoming more efficient, \method{} can also discover new and effective behaviors. For instance, in the plug insertion task, we observe that the policy sometimes opts for a behavior not seen in any of the insertion demos: grabbing the connector when lying on the side, rather than upright, and reorienting it in-air to do the insertion (see Figure~\ref{fig:appendix_lift_types} and see supplementary video). This validates the importance of the RL component in \method{} allowing policies to go beyond the demonstrations and produce more efficient and robust solutions.
In other tasks, we also observe emerging behaviors. For instance, the plug lifting policies use more robust grasps that the ones in the demonstrations which mostly used the fingertips for holding the plug. In the nut and bolt task, we see a more intentional and efficient use of a single finger the rotate the nut and finish the screwing. For cube reorientation, the behavior are clearly different from the demonstrations as the two demonstrations provided only contain unstructured interaction with the connector rather than an intentional re-orientation. 

\begin{figure}
    \centering
    \includegraphics[height=4cm]{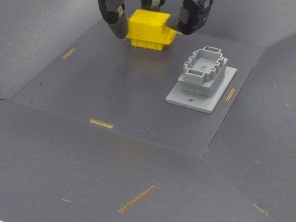}
    \includegraphics[height=4cm]{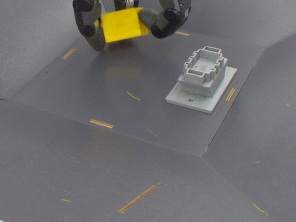}
    \caption{Example of the two lifting strategies used by the plug insertion policy. The one on the left matches the grasp used when generating demonstrations in simulation. The one on the right is a new grasping solution that \method{} learned with RL.}
    \vspace{-0.2cm}
    \label{fig:appendix_lift_types}
\end{figure}

\subsubsection{ Behavioral differences between \method{} and the baselines}

In Section~\ref{sec:results}, we showed that the baselines cannot match the performance of the distilled policies from \method{}.
Qualitatively we observe that SAC-X policies result in more jittery but also more diverse behaviors. As a result, it becomes hard to distill these policies which results in lower performance even in simulation for the task of plug insertion.

With regards to the demonstrations collected on the real setup, we observed smoother behaviors but still large diversity in the behaviors collected. This is reasonable as each person has its own preferences towards how to perform the task during teleoperation. For instance, some teleoperators opt for  lifting and subtly throwing the connector to reorient it while other perform the reorientation in-hand. As a result, the learned policies struggle to capture all this behaviors into a single policy resulting in more jerky behaviors than the distilled policies from \method{}. We believe this explains why it is hard to learn from our teleoperation data and suggest that either more demonstrations would be needed to further boost performance or that teleoperators would need to receive clear instructions on how to perform the task to make demonstrations as similar as possible to ease learning.

\subsubsection{Failure modes during sim-to-real transfer}

In simulation, \method{} provides policies that solve the tasks reliably and with minimal failure. However, when we distill and transfer these policies to the real setup, we observe some failure modes. For cube reorientation, the three failures during the evaluation (Section~\ref{table:real_performance}) come from the cube ending in one corner of the basket and the policy struggling to bring it to center with the right orientation on time. For plug lifting, the main failure mode is failing to hold the object reliably as small finger motions can break the grasp.

The task of plug insertion is the hardest to transfer zero-shot from simulation. While our policies manage to reorient and successfully grasp the plug, they sometimes fail when the plug and socket are aligned and a precise insertion behaviour is needed. This results in episodes with multiple insertion attempts. For the baseline policies, this presents two additional failure modes:  understanding the orientation of the connector and failing at recognizing that the plug wasn't correctly inserted. As a result, when executing the baseline policies the robot sometimes attempts to insert the plug in the wrong orientation or leaves the plug on top of the socket rather than finalizing the insertion. The SAC-X baseline also struggles with reliably lifting the plug.

\subsubsection{Lifting generalization}

We found that the policy for plug lifting is also capable of lifting objects of different shapes and colors, like a cube or a lemon, with reasonable success rate. While generalization is beyond the scope of this work, it is promising to see that the policy can already adapt to these new objects which present different challenges. For instance, the lemon tends to roll in the basket, making it much harder to lift, while the cube is of a different color and shape resulting in clear visual domain gap with respect to the plug (see videos of lifting these objects in the supplementary material).

\end{document}